**CNN-Based Framework for Pedestrian Age and Gender Classification Using Far-View Surveillance in Mixed-Traffic Intersections**

**Shisir Shahriar Arif**
Graduate Research Assistant, Department of Civil Engineering
Bangladesh University of Engineering and Technology (BUET), Dhaka, Bangladesh, 1000
Tel: +880-2-55167100, Ext. 7225; Fax: +880-2-58613046; Email: shahriarshisir421@gmail.com

**Md. Muhtashim Shahrier**
Graduate Research Assistant, Department of Civil Engineering
Bangladesh University of Engineering and Technology (BUET), Dhaka, Bangladesh, 1000
Tel: +880-2-55167100, Ext. 7225; Fax: +880-2-58613046; Email: shahriermuhtashim@gmail.com

**Nazmul Haque**
Lecturer, Accident Research Institute (ARI)
Bangladesh University of Engineering and Technology (BUET), Dhaka, Bangladesh, 1000
Tel: +880-2-55167100, Ext. 7897; Fax: +880-2-58613046; Email: nhaque@ari.buet.ac.bd

**Md Asif Raihan***
Associate Professor, Accident Research Institute (ARI)
Bangladesh University of Engineering and Technology (BUET), Dhaka, Bangladesh, 1000
Cell: + 8801911142802; Email: raihan@ari.buet.ac.bd

**Md. Hadiuzzaman**
Professor, Department of Civil Engineering
Bangladesh University of Engineering and Technology (BUET), Dhaka-1000
Tel: +880-2-55167100, Ext. 7225; Fax: +880-2-58613046; Email: mhadiuzzaman@ce.buet.ac.bd

Word count: 5,972 words text + 6*250 (6 table) = 7,472 words

**ABSTRACT**
Pedestrian safety remains a pressing concern in congested urban intersections, particularly in low- and middle-income countries where traffic is multimodal, and infrastructure often lacks formal control. Demographic factors like age and gender significantly influence pedestrian vulnerability, yet real-time monitoring systems rarely capture this information. To address this gap, this study proposes a deep learning framework that classifies pedestrian age group and gender from far-view intersection footage using convolutional neural networks (CNNs), without relying on facial recognition or high-resolution imagery. The classification is structured as a unified six-class problem, distinguishing adult, teenager, and child pedestrians for both males and females, based on full-body visual cues. Video data was collected from three high-risk intersections in Dhaka, Bangladesh. Two CNN architectures were implemented: ResNet50, a deep convolutional neural network pretrained on ImageNet, and a custom lightweight CNN optimized for computational efficiency. Eight model variants explored combinations of pooling strategies and optimizers. ResNet50 with Max Pooling and SGD achieved the highest accuracy (86.19%), while the custom CNN performed comparably (84.15%) with fewer parameters and faster training. The model's efficient design enables real-time inference on standard surveillance feeds. For practitioners, this system provides a scalable, cost-effective tool to monitor pedestrian demographics at intersections using existing camera infrastructure. Its outputs can shape intersection design, optimize signal timing, and enable targeted safety interventions for vulnerable groups such as children or the elderly. By offering demographic insights often missing in conventional traffic data, the framework supports more inclusive, data-driven planning in mixed-traffic environments.

**INTRODUCTION**

Pedestrians are among the most vulnerable road users, facing risks in increasingly congested and complex urban environments. Intersections, in particular, act as key conflict zones where motorized vehicles, non-motorized modes, and pedestrian flows interact in shared spaces that are often chaotic. Understanding pedestrian behavior and characteristics, particularly age group and gender, is crucial for improving safety, as these factors influence walking speed, crossing decisions, and overall vulnerability within traffic environments (*1*). Studies have shown that age-related mobility differences influence crossing speed, risk perception, and pedestrian-vehicle conflicts (*2*, *3*). Elderly pedestrians are at higher risk due to slower walking speeds, reduced cognitive processing abilities, and impaired gap perception and they often require longer crossing times at intersections (*4*). On the other hand, children have limited situational awareness and impulsive behaviors, making them particularly vulnerable in unregulated mixed-traffic environments (*3*). Gender-based differences in pedestrian behavior further affect exposure to risk. Studies indicate that males tend to engage in riskier crossing behavior, making faster but less cautious crossing decisions (*5*). Conversely, females exhibit greater compliance with pedestrian signals but often express higher safety concerns, influencing their choice of crossing locations (*6*). Despite their importance, these demographic factors are rarely integrated into real-time monitoring systems, traffic modeling, or infrastructure design processes.

In many densely populated urban areas of South Asia and other low- and middle-income countries, traffic systems involve diverse modes such as buses, rickshaws, motorcycles, and bicycles operating alongside large pedestrian populations. The lack of formal infrastructure and unpredictable traffic patterns increase pedestrian risks, especially at intersections (*7*). Yet conventional traffic studies in these contexts often overlook critical demographic insights. Age and gender are rarely integrated into safety planning. This limits the ability of transportation agencies to design inclusive, data-driven interventions that address the specific vulnerabilities of different pedestrian groups.

While Convolutional Neural Network (CNN)-based models have been widely applied to various traffic-related tasks, such as vehicle detection and general object recognition, their use in identifying pedestrian demographic attributes, specifically age group and gender, within complex and mixed-traffic urban environments remains notably limited (*8*, *9*). This gap is particularly evident in developing countries, where unique traffic conditions such as high pedestrian density, diverse vehicle types, and irregular traffic movements create additional challenges for accurate and reliable detection systems. Traditional pedestrian data collection techniques including manual observation, LiDAR tracking, and video surveillance focus primarily on movement analysis but often fail to capture demographic characteristics such as age and gender (*10*, *11*). Manual observations are labor-intensive and subjective, while sensor-based methods such as RFID and LiDAR detect pedestrian movement but cannot infer demographic information (*11*).

Despite advancements, many standard CNN architectures rely heavily on facial features for demographic classification, which is infeasible in low-resolution surveillance footage where facial attributes are blurred or occluded. To address this, custom CNN architectures have been developed to focus on body-based attributes, such as height, gait, and clothing features, for pedestrian age and gender classification. Levi and Hassner proposed a CNN optimized for simultaneous age and gender classification, improving its adaptability to real-world surveillance settings with varying image quality (*12*). Similarly, Chowdhury et al. developed a CNN specifically trained on far-view pedestrian images, demonstrating that body proportions and movement patterns can serve as reliable demographic indicators when facial recognition is unavailable, compensating for missing facial details and improving classification reliability (*13*). Recent advancements in multi-task learning and attention mechanisms have further improved CNN-based pedestrian classification. Wu et al. implemented attention-based CNNs, which dynamically focus on relevant pedestrian attributes while ignoring background distractions, improving classification accuracy in partially occluded and crowded urban environments (*14*). Hybrid architectures that integrate CNN-based feature extraction with transformer models have also been explored, enhancing long-range feature dependencies and improving classification in complex traffic settings (*15*).

One of the key challenges in traffic surveillance is low-resolution pedestrian detection in far-view cameras. JCS-Net, developed by Pang et al., integrated super-resolution techniques, improving classification accuracy in small-scale pedestrian images (*16*). Occlusions and dynamic pedestrian movement patterns further complicate classification in mixed-traffic settings. Zhu et al. proposed a multi-attribute pedestrian classification framework, integrating posture, movement, and clothing cues, which enhanced demographic classification accuracy in dense urban areas with diverse pedestrian behaviors (*17*).

This study proposes a CNN-based classification framework designed to detect both age group and gender of pedestrians from far-view traffic surveillance footage in Dhaka's mixed-traffic intersections. Unlike most existing methods that classify age and gender separately or rely heavily on close-range facial features, this approach treats demographic classification as a single, six-class problem using full-body visual cues such as posture, clothing, and body proportions. The model was developed using real-world video footage collected from three four-legged intersections in Dhaka, Motsho Bhaban, Banglamotor, and Bijoy Sarani which are selected based on pedestrian crash prevalence. More than 3,800 frames were manually annotated and processed to create a custom dataset of over 30,000 pedestrian images, balanced across six demographic classes using data augmentation. Two CNN architectures were implemented: a lightweight custom CNN designed for computational efficiency and ResNet50 as a deep-learning benchmark. Both were trained and evaluated using carefully controlled variations in pooling strategies such as Global Average Pooling (GAP) and Max Pooling (MP), and optimization algorithms such as Adaptive Moment Estimation (Adam) and Stochastic Gradient Descent (SGD) to assess trade-offs between accuracy and model complexity. By integrating age and gender classification into intersection-level surveillance, the proposed framework offers practical value for pedestrian safety assessment, signal timing adjustment, and inclusive infrastructure planning. Its ability to operate on far-view low resolution images and lightweight architecture makes it deployable on low-cost surveillance systems enabling demographic-aware traffic monitoring without requiring facial recognition or high-end computing infrastructure.

The remainder of the paper is organized as follows. The Methodology section outlines the design of the classification framework, including model architecture and the formulation of the six-class problem. The Data Collection and Preparation section describes the study sites, video recording process, and annotation techniques used to generate the pedestrian image dataset. The Model Development and Training section details the implementation of ResNet50 and the custom CNN. The Result and Discussion section presents the performance comparison across model variants using evaluation metrics such as accuracy, precision, recall, F1-score, and PR-AUC. Finally, the Conclusion summarizes key findings and discusses implications for practice and future research.

## METHODOLOGY

This study presents a computer vision–based framework to classify pedestrian age group and gender using far-view video footage from urban intersections. The classification task was structured as a unified six-class problem: Male Adult, Male Teenager, Male Child, Female Adult, Female Teenager, and Female Child. Rather than treating age and gender as separate binary classifications, this design integrates them into a single output space, allowing for joint inference that reflects real-world decision-making needs.

Two convolutional neural network (CNN) architectures were selected for training and evaluation. The first was ResNet50, developed by He et al., a 50-layer deep residual network pretrained on the ImageNet dataset (*18*). It was chosen for its strong performance on low-resolution and occluded imagery. It is made possible by its skip connections that enable stable training of deep architectures. ResNet50 is widely used in image classification tasks due to its ability to learn generalizable visual patterns even under noisy or constrained visual conditions, qualities critical for far-view surveillance environments (*18*). The final fully connected layer was replaced with a six-node softmax layer to match the classification task.

A lightweight custom CNN was designed in parallel specifically for this application. The custom model offered a shallower architecture and significantly fewer parameters, aiming to reduce training time and enable deployment on embedded systems with limited computing resources. While ResNet50 offered strong out of the box generalization, the custom CNN provided flexibility and computational efficiency.

*ResNet50 Architecture*

The first model architecture evaluated was ResNet50, a 50-layer deep residual convolutional neural network originally trained on the ImageNet dataset. It was selected due to its strong generalization capabilities under challenging visual conditions, such as low resolution, partial occlusion, and cluttered backgrounds which are common in far-view intersection footage. ResNet50 is particularly effective in these contexts because of its residual learning design, which includes identity-based skip connections that enable the stable training of deep networks by alleviating the vanishing gradient problem (*18*).

For this study, the base ResNet50 architecture was retained, and only the final classification layer was modified. The original 1000-class fully connected (FC) output layer was removed and replaced with a dense layer of six neurons followed by a dropout layer and a softmax activation function, matching the six pedestrian demographic classes used in this study. The model was initialized with pretrained ImageNet weights to leverage general visual features and accelerate convergence. Two pooling strategies, Global Average Pooling (GAP) and Max Pooling (MP) were tested after the final convolutional block to evaluate their effect on generalization and classification performance.

ResNet50 models were trained in two phases. In the first phase, the convolutional base was frozen and only the top layers were trained on the pedestrian dataset. In the second phase, the top 100 layers of the base network were unfrozen and fine-tuned using a reduced learning rate to adapt higher-level features to the target domain. This two-stage process allowed for both efficient learning and deeper domain adaptation without destabilizing the pretrained weights (**Figure 1**).

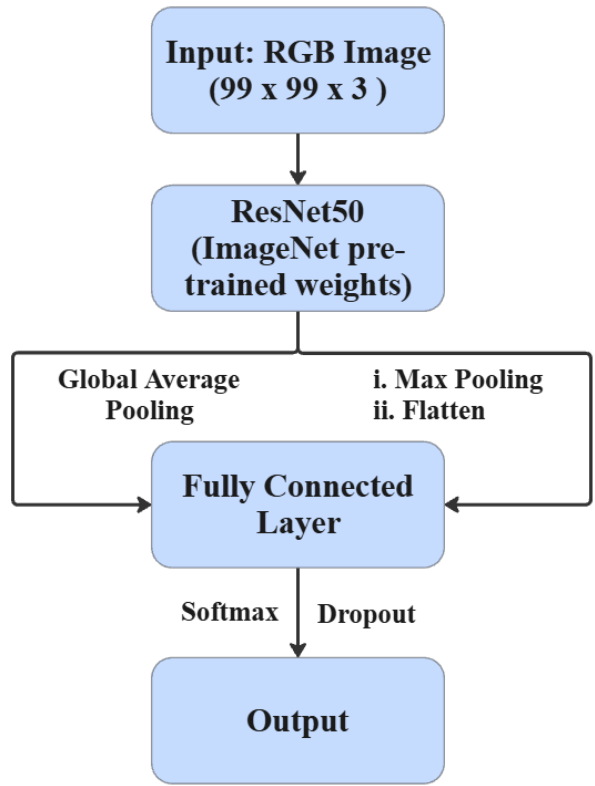

**Figure 1 ResNet50 Architecture**

*Custom CNN Architecture*

The custom architecture (**Figure 2**) was designed from scratch using a modular configuration that balances classification capacity with computational efficiency. The network accepts RGB pedestrian images resized to 99×99 pixels, preserving sufficient spatial detail for capturing body-based visual cues while maintaining manageable input dimensions suitable for real-time inference.

The model consists of four sequential convolutional blocks, each designed to extract progressively abstract and high-level features. The first block employs 32 convolutional filters of size 3×3 with padding set to 'same' to preserve spatial resolution, followed by batch normalization, a ReLU (Rectified Linear Unit) activation function, and a 2×2 max pooling layer to reduce dimensionality. The second block increases the filter count to 64 while maintaining the same configuration, again followed by batch normalization, ReLU, and max pooling. The third and fourth blocks further expand the network's representational depth with 128 and 256 filters respectively, each using 3×3 kernels and the same post-processing operations. This hierarchical design allows the network to extract increasingly complex spatial features from the input images.

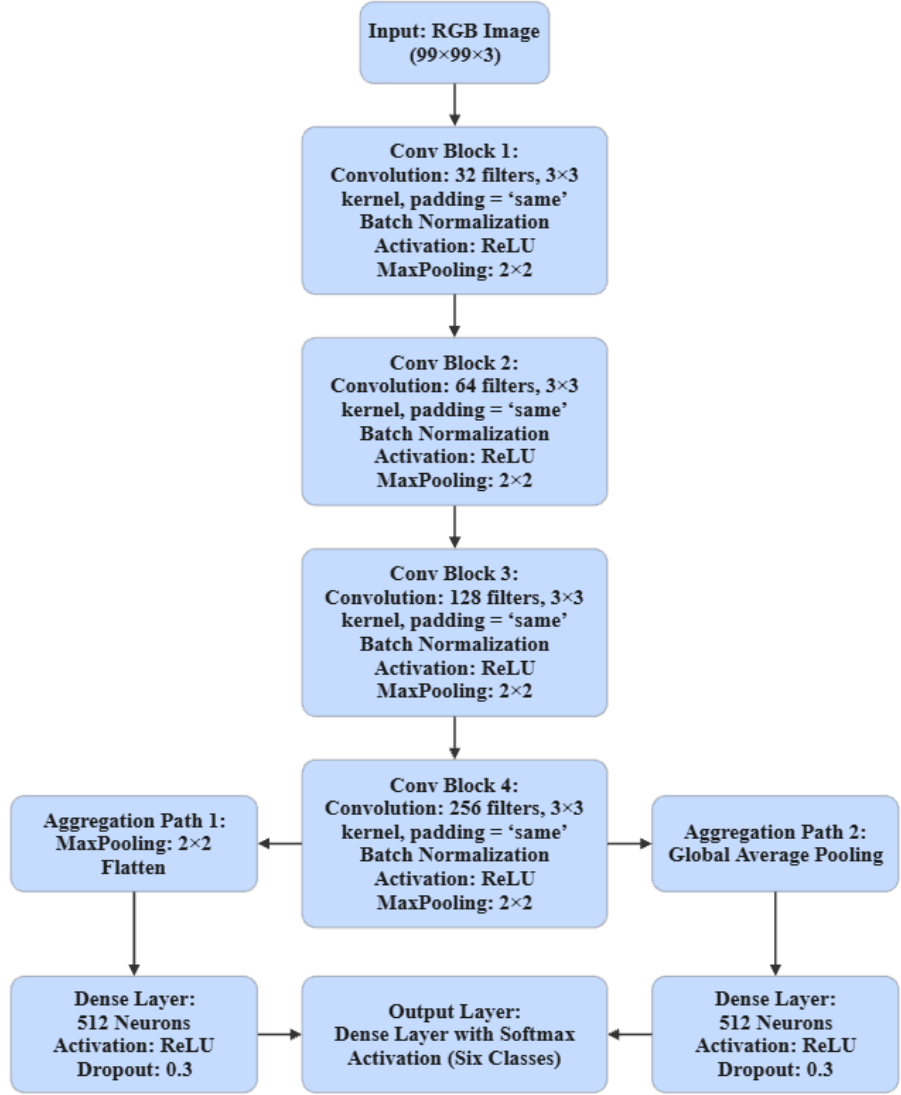

**Figure 2 Custom CNN Architecture**

After the final convolutional block, two alternative strategies were explored for aggregating the extracted feature maps. In the first variant, an additional 2×2 max pooling operation was applied, followed by flattening the feature maps into a one-dimensional vector. In the second variant, global average pooling (GAP) was employed to compute the spatial mean of each feature map, resulting in a more compact feature representation. Both approaches were followed by a fully connected dense layer with 512 neurons activated by ReLU, along with a dropout layer (dropout rate = 0.3) to mitigate overfitting.

The final classification layer consists of six neurons with softmax activation, producing a probability distribution over the six defined demographic classes. Depending on the pooling strategy used, the total number of trainable parameters in the model ranged from approximately 524,000 to 1.57 million, making it significantly more lightweight than standard deep architectures such as ResNet50. The compact size of the custom CNN allowed fast training and made it well-suited for deployment in low bandwidth and power constrained environments. Despite its relatively simple structure, the custom model delivered competitive performance across evaluation metrics, with particularly strong results observed when using max pooling in combination with stochastic gradient descent (SGD) optimization.

*Hyperparameter Tuning and Experimental Design*

In order to ensure a fair and systematic comparison between the ResNet50 and custom CNN models, a set of common and varied hyperparameters was established. Several hyperparameters were kept constant across all models to isolate the effects of optimization strategies and pooling choices. These common hyperparameters included an input image size of 99×99 pixels which was selected to retain key pedestrian body cues while maintaining manageable computational load. The batch size was fixed at 8, balancing gradient stability with efficient GPU memory usage. A dropout rate of 0.3 was applied after the dense layer in all models to mitigate overfitting. It was particularly important given the high visual similarity among certain pedestrian classes. All models were trained to classify into six demographic categories using a softmax output layer with categorical cross-entropy loss.

The training process spanned up to 70 epochs with early stopping implemented. This safeguard prevented overfitting while reducing unnecessary computation. For ResNet50-based models, a two-stage training strategy was adopted. Initially, the base convolutional layers pretrained on ImageNet were frozen, and only the newly added top layers were trained to learn task-specific features. This phase leveraged general-purpose visual features while reducing training time. In the second stage with additional 30 epochs, the top 100 layers of ResNet50 were unfrozen and fine-tuned using a lower learning rate, allowing the model to adapt high-level features to the domain of far-view pedestrian images without destabilizing the pretrained weights.

To evaluate the impact of optimizer choice and pooling strategy, four variants were trained for each architecture, combining two optimizers (Adam and SGD with momentum) with two fully connected layer pooling strategies (Global Average Pooling and Max Pooling). Adam was initialized with a learning rate of 0.0001, while SGD used 0.01 with 0.9 momentum during initial training. During fine-tuning for ResNet50 models, these values were reduced to 0.00001 for Adam and 0.001 for SGD to stabilize convergence.

Pooling strategies were varied to assess their effect on model capacity and overfitting. Global Average Pooling (GAP) reduces each feature map to a single scalar by computing its spatial mean, thereby minimizing parameters and discouraging overfitting. In contrast, Max Pooling (MP) preserves the most prominent local activations within each region and, when followed by flattening, enables the dense layers to learn from spatially detailed features. Although this often increases the number of parameters and the risk of overfitting. Max Pooling was popularized in early architectures like AlexNet for its ability to enhance translation invariance and reduce spatial dimensions efficiently (*19*), while Global Average Pooling (GAP), introduced by (*20*), emerged as a parameter-efficient alternative that promotes better generalization.

The choice to explore both pooling and optimizer combinations stemmed from the visual challenges posed by far-view images captured in uncontrolled intersection environments characterized by low resolution, clutter, and body-based visual cues rather than facial features. By designing eight model variants across both architectures (ResNet50 and custom CNN), the experimental framework aimed to identify combinations that balance performance with computational efficiency.

The dataset was divided into training (70%), validation (20%), and test (10%) sets to assess classifier performance. Stratified sampling was used to preserve class distribution. Models were evaluated using multiclass classification metrics suited to imbalanced datasets, including accuracy, precision, recall,

F1-score, confusion matrix, precision–recall (PR) curves, and area under the PR curve (PR-AUC). These metrics helped evaluate the model thoroughly, especially for pedestrian classes that had fewer examples. They also guided the final model choice and helped understand where the model struggled.

## DATA COLLECTION AND PREPARATION

To ensure contextual relevance and safety-critical insights, video data were collected at three four-legged intersections in Dhaka, Bangladesh—Motsho Bhaban, Banglamotor, and Bijoy Sarani. These locations were chosen based on pedestrian crash concentration data from the Accident Research Institute (ARI) between 2016 and 2020. The selected intersections not only experienced the highest number of crashes involving pedestrians but also exhibited a significant share of fatal and grievous injury crashes (**Figure 3, Figure 4**). The presence of diverse road users, inadequate pedestrian infrastructure, and complex geometric layouts made these intersections ideal for capturing rich pedestrian behavior under mixed traffic conditions.

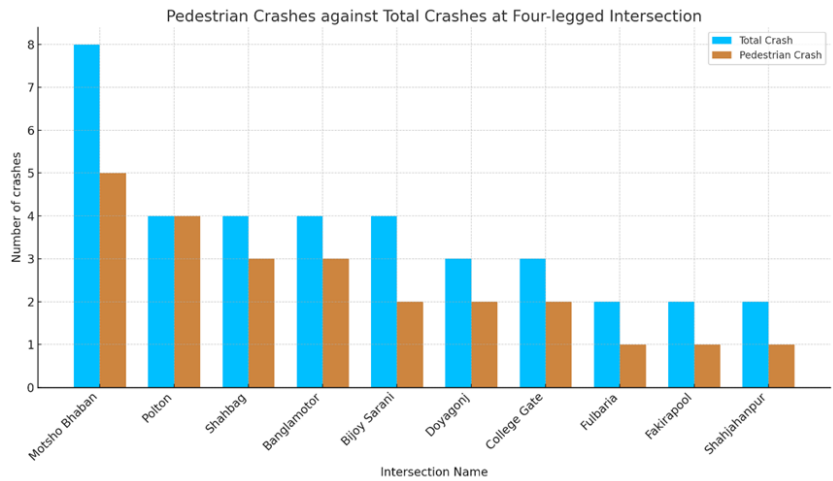

**Figure 3 Pedestrian Crashes against Total Crashes at Four-legged Intersections**

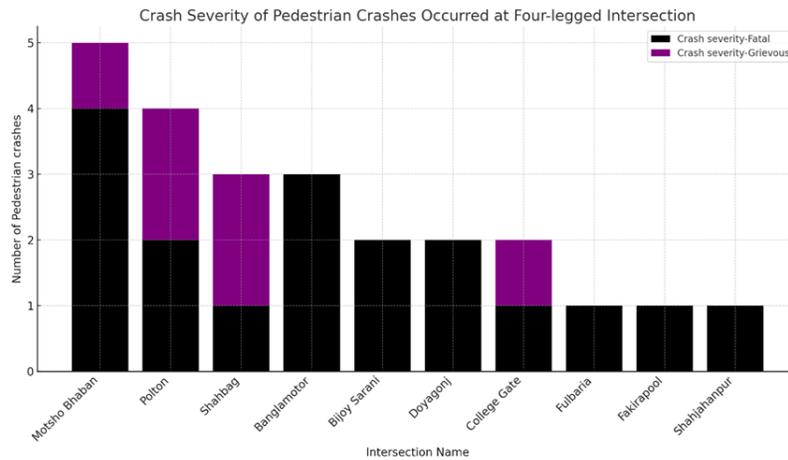

**Figure 4 Crash Severity of Pedestrian Crashes Occurred at Four-legged Intersection**

Two high-resolution camcorders, Panasonic HC-PV100GW and Sony FDR-AX700 were deployed to record traffic activity. Camera positioning was carefully calibrated to ensure optimal visibility, typically at heights between 20 and 50 feet. The field of view was adjusted to cover at least two legs of each intersection while minimizing occlusion from adjacent vehicles. The camera angle was configured to exclude horizon lines and capture stable, side-profile views of pedestrians.

Due to lighting constraints affecting image quality in afternoon hours, video footage was collected only during morning peak periods (8 AM to 12 PM), when traffic volumes were high, and lighting conditions were consistent. The effective total recording duration amounted to five hours—two hours each at Banglamotor and Bijoy Sarani, and one hour at Motsho Bhaban.

Video frames were extracted at 5-second intervals to build a robust image dataset while avoiding redundancy, striking a balance between pedestrian coverage and computational efficiency. This interval yielded 3,871 frames from the five hours of footage, with each frame saved at a resolution of 1920×1080 pixels at 25 frames per second (fps).

Extracted frames were uploaded to Roboflow, where each pedestrian was manually annotated with bounding boxes and assigned to one of six predefined demographic classes: Male Adult, Female Adult, Male Teenager, Female Teenager, Male Child, and Female Child (**Figure 5**). The annotation process ensured that the dataset captured relevant demographic information for use in deep learning-based classification. After annotations were completed, the dataset was downloaded in COCO JSON format without applying any modifications such as resizing, augmentation, or pre-splitting, preserving the raw annotations for further processing. A custom Python script was then used to crop each pedestrian bounding box from the annotated frames, with the resulting images saved into separate folders based on their class labels, thereby retaining only the regions of interest and excluding unnecessary background elements.

The raw dataset, following cropping, was highly imbalanced. For example, the Male Adult class contained 35,806 images, whereas the Female Teenager, Male Child, and Female Child categories contained fewer than 350 images each. Such class disparities can severely skew deep learning models toward the majority class, degrading performance on underrepresented groups. However, balancing was not applied immediately. Instead, the full imbalanced dataset was first split into three subsets for model development.

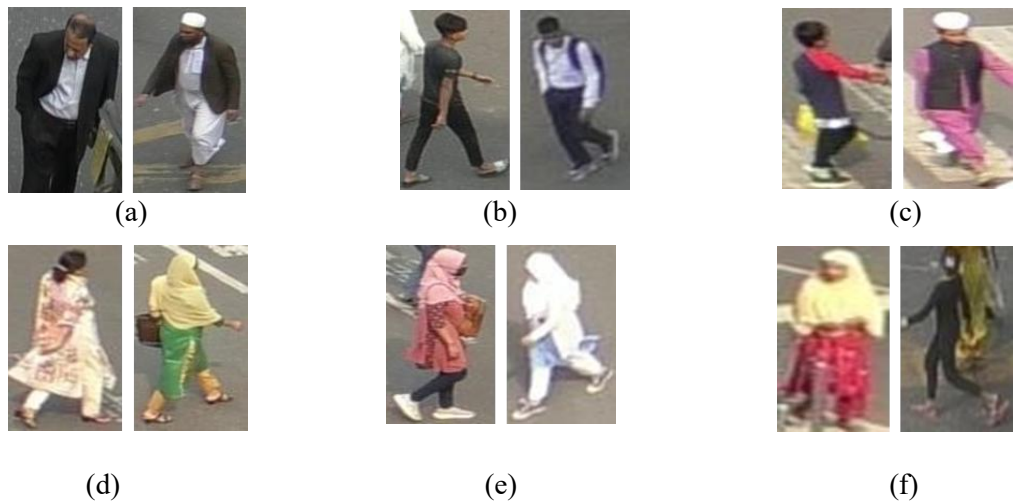

**Figure 5 Sample Images from (a) Male Adult Class (b) Male Teenager Class (c) Male Child Class (d) Female Adult Class (e) Female Teenager Class (f) Female Child Class**

The dataset was partitioned into training, validation, and test sets in a 70:20:10 ratio. This split was performed class-wise to ensure that each subset retained the original distribution. The training set was used to update model parameters, the validation set guided hyperparameter tuning and monitored generalization, and the test set was strictly reserved for final evaluation. The imbalance was preserved in the validation and test sets to reflect real-world demographic frequencies during model assessment. Only the training set underwent balancing operations.

To mitigate bias during learning, the training set was restructured through two methods. First, the Male Adult class was randomly down sampled to 5,000 images. Then, data augmentation was applied to all underrepresented classes until each contained 5,000 images. These steps ensured that each demographic category contributed equally to the model's learning process, reducing the risk of overfitting to dominant visual patterns. Data augmentation was applied only to the training set using TensorFlow's preprocessing pipeline, introducing variations such as flips, rotations, shifts, shears, and zooms to enhance model generalization, particularly in far-view footage where clothing and posture are key cues. Validation and test sets remained unaltered to ensure reliable evaluation on real images. Despite augmentation, classification challenges persisted for underrepresented classes like Female Teenager and Male Child, likely due to overlapping body proportions and limited distinguishing features. The final training set was balanced with 5,000 images per class, providing a solid base for training and comparing deep CNN models.

**MODEL DEVELOPMENT AND TRAINING**

Two convolutional neural network (CNN) architectures were developed and evaluated. The first was ResNet50, a deep residual network pretrained on ImageNet, selected for its proven robustness in noisy and low-resolution conditions. The second was a custom lightweight CNN, designed from scratch for fast training, low parameter count, and potential deployment in resource constrained environments.

Each architecture was evaluated using two different pooling strategies, Global Average Pooling (GAP) and Max Pooling (MP), as well as two optimizers, Adam and Stochastic Gradient Descent (SGD) with momentum. These choices resulted in a total of eight model variants, designed to explore the trade-offs between generalization, convergence speed, and computational efficiency.

All models were trained using a Tesla P100 GPU on Kaggle's cloud platform, ensuring efficient training for both deep and lightweight architectures.

The different combinations of pooling layers and optimizers used in ResNet50 and custom CNNs are summarized in **Table 1**.

**TABLE 1 Model Comparison Table Based on Varied Hyperparameters**

| Model Name | Architecture | Pooling Layer | Weights | Optimizer | Learning Rate (Initial) | Learning Rate (Fine-Tune) | Total Parameters/ Trainable parameters |
|---|---|---|---|---|---|---|---|
| Model 1 | ResNet50 | GAP | ImageNet | Adam | 0.0001 | 0.00001 | 24,639,878/ 1,052,166 |
| Model 2 | ResNet50 | MP | ImageNet | Adam | 0.0001 | 0.00001 | 27,785,606/ 4,197,894 |
| Model 3 | ResNet50 | GAP | ImageNet | SGD | 0.01 | 0.001 | 24,639,878/ 1,052,166 |
| Model 4 | ResNet50 | MP | ImageNet | SGD | 0.01 | 0.001 | 27,785,606/ 4,197,894 |
| Model 5 | Custom CNN | GAP | None | Adam | 0.00001 | None | 524,998/ 524,038 |
| Model 6 | Custom CNN | MP | None | Adam | 0.00001 | None | 1,573,574/ 1,572,614 |
| Model 7 | Custom CNN | GAP | None | SGD | 0.001 | None | 524,998/ 524,038 |
| Model 8 | Custom CNN | MP | None | SGD | 0.001 | None | 1,573,574/ 1,572,614 |

**RESULTS AND DISCUSSION**

The study evaluates the performance of eight models in distinguishing age and gender categories under mixed-traffic conditions. It explores the training, validation, and testing performance of ResNet50-based and Custom CNN architectures, examining their accuracy, precision, recall, F1-score, PR curve, PR-AUC score and confusion matrix.

Accuracy quantified the proportion of total correct predictions out of all predictions. It is calculated by **Equation 1**.

$$Accuracy = \frac{TP + TN}{TP + TN + FP + FN} \tag{1}$$

Precision measures how many of the instances predicted as positive were actually correct. It is calculated by **Equation 2**.

$$Precision = \frac{TP}{TP + FP} \tag{2}$$

Recall indicates how many of the actual positives were correctly identified by the model. It is calculated by **Equation 3**.

$$Recall = \frac{TP}{TP + FN} \tag{3}$$

F1-Score is a weighted average of precision and recall. It considers both false positives and false negatives. It is calculated by **Equation 4.**

$$F1 - score = 2 \times \frac{Precision \times Recall}{Precision + Recall} \tag{4}$$

To account for the imbalance across classes, both macro and weighted averages were computed. Macro averaging treated each class equally, regardless of size. It is calculated by **Equation 5.**

$$Macro - average = \frac{1}{N} \times \sum_{i=1}^{N}(Metric_i) \tag{5}$$

Weighted averaging adjusted each class contribution based on its sample count. It is calculated by **Equation 6.**

$$Weighted - average = \frac{1}{\sum_{i=1}^{N} S_i} \times \sum_{i=1}^{N} S_i \times Metric_i \tag{6}$$

*Accuracy*

Among all models, ResNet50 with Max Pooling and SGD (Model 4) achieved the highest test accuracy at 86.19%, closely followed by ResNet50 with Global Average Pooling and SGD (Model 3) at 84.51%. The best-performing custom model, Model 8 (Custom CNN with Max Pooling and SGD), achieved a test accuracy of 84.15%. Notably, Model 8 used only 1.57 million trainable parameters which is less than half of the most complex ResNet50 models, demonstrating a favorable trade-off between performance and computational efficiency. The training duration also reflects this contrast as Model 8 converged in just 18 epochs, whereas ResNet50-based models required up to 53 epochs (**Table 2**).

The observed performance differences can be attributed with the architectural depth and weight initialization strategy. ResNet50 benefited from pretraining on ImageNet, enabling strong feature transfer even in far-view pedestrian imagery. However, the custom CNN demonstrated robustness with far fewer parameters, indicating that shallow architectures can still be effective when paired with appropriate regularization and pooling strategies.

*Arif, Shahrier, Haque, Raihan, and Hadiuzzaman*

**TABLE 2 Validation and Testing Accuracy and Loss across Different Models**

| Model Name | Validation Accuracy | Validation Loss | Testing Accuracy | Total Epochs | Parameters Trained |
|---|---|---|---|---|---|
| Model 1 | 0.8218 | 0.5531 | 0.8238 | 28 | 1,052,166 |
| Model 2 | 0.8316 | 0.5289 | 0.8279 | 53 | 4,197,894 |
| Model 3 | 0.8323 | 0.5672 | 0.8451 | 40 | 1,052,166 |
| Model 4 | 0.8698 | 0.4580 | 0.8619 | 36 | 4,197,894 |
| Model 5 | 0.8257 | 0.5517 | 0.8320 | 28 | 524,038 |
| Model 6 | 0.8005 | 0.5812 | 0.8102 | 28 | 1,572,614 |
| Model 7 | 0.8286 | 0.5592 | 0.8265 | 19 | 524,038 |
| Model 8 | 0.8380 | 0.5325 | 0.8415 | 18 | 1,572,614 |

*Class wise Precision, Recall and F1 score*

Precision and recall values further illuminate the class-wise strengths and weaknesses of each model. All models consistently performed best on the Male Adult class, achieving precision above 90% and recall up to 94% in Model 4 (**Table 3, Table 4**). This is unsurprising given the large number of training instances and the relatively distinctive features such as height, build, and clothing style associated with adult males. Female Adult also exhibited relatively high precision and recall across models due to similar reasons.

**TABLE 3 Precision Comparison Table**

| Model | Female Adult | Female Child | Female Teenager | Male Adult | Male Child | Male Teenager | Macro Avg | Weighted Avg |
|---|---|---|---|---|---|---|---|---|
| Model 1 | 0.7489 | 0.4545 | 0.25 | 0.9022 | 0.1111 | 0.4619 | 0.4881 | 0.8155 |
| Model 2 | 0.7413 | 0.4615 | 0.2353 | 0.9172 | 0.1579 | 0.4762 | 0.4982 | 0.8255 |
| Model 3 | 0.7455 | 0.6 | 0.4 | 0.9396 | 0.4 | 0.5086 | 0.5899 | 0.8505 |
| Model 4 | 0.7713 | 0.6471 | 0.2857 | 0.9107 | 0.8571 | 0.6538 | 0.6876 | 0.8509 |
| Model 5 | 0.7166 | 0.3333 | 0.2143 | 0.899 | 0.6667 | 0.5846 | 0.5691 | 0.8195 |
| Model 6 | 0.6761 | 0.4167 | 0.25 | 0.9163 | 0.4167 | 0.503 | 0.5298 | 0.8155 |
| Model 7 | 0.7028 | 0.6 | 0.2 | 0.8949 | 0.4 | 0.619 | 0.5695 | 0.8161 |
| Model 8 | 0.747 | 0.5 | 0.3333 | 0.9073 | 0.3333 | 0.6 | 0.5702 | 0.8325 |

**TABLE 4 Recall Comparison Table**

| Model | Female Adult | Female Child | Female Teenager | Male Adult | Male Child | Male Teenager | Macro Avg | Weighted Avg |
|---|---|---|---|---|---|---|---|---|
| Model 1 | 0.7331 | 0.1923 | 0.1034 | 0.9227 | 0.08 | 0.5064 | 0.4229 | 0.8238 |
| Model 2 | 0.8136 | 0.2308 | 0.1379 | 0.9007 | 0.12 | 0.5128 | 0.4526 | 0.8279 |
| Model 3 | 0.8750 | 0.4615 | 0.2069 | 0.8920 | 0.32 | 0.5705 | 0.5543 | 0.8451 |
| Model 4 | 0.8432 | 0.4231 | 0.0689 | 0.9453 | 0.24 | 0.4359 | 0.4927 | 0.8619 |
| Model 5 | 0.7500 | 0.1538 | 0.1034 | 0.9320 | 0.08 | 0.4872 | 0.4177 | 0.8319 |
| Model 6 | 0.8093 | 0.3846 | 0.1379 | 0.8687 | 0.20 | 0.5449 | 0.4909 | 0.8102 |
| Model 7 | 0.7415 | 0.2308 | 0.2069 | 0.9307 | 0.08 | 0.4167 | 0.4344 | 0.8265 |
| Model 8 | 0.7945 | 0.4231 | 0.1379 | 0.9260 | 0.28 | 0.4615 | 0.5038 | 0.8415 |

In contrast, minority classes such as Female Child and Female Teenager were much harder to classify. Even the best-performing models struggled with these categories: precision and recall values often dipped below 40%, with F1-scores as low as 20% in certain configurations. These results underscore the limitations imposed by data imbalance and visual similarity across adjacent age groups. For instance, teenagers and children may exhibit overlapping body proportions and dress styles, particularly in far-view imagery where facial cues are unavailable.

F1-score analysis reveals similar trends. While majority classes maintain high F1-scores, underrepresented classes exhibit volatile and often poor values, reflecting instability in the model's ability to generalize. The highest macro-averaged F1-score among all models was 0.568 (Model 3), whereas Model 8 achieved 0.528. Though these values are moderate, the weighted average F1-scores were consistently higher due to dominance by better-performing majority classes (**Table 5**).

**TABLE 5 F1-score Comparison Table**

| Model | Female Adult | Female Child | Female Teenager | Male Adult | Male Child | Male Teenager | Macro Avg | Weighted Avg |
|---|---|---|---|---|---|---|---|---|
| Model 1 | 0.7409 | 0.2703 | 0.1463 | 0.9123 | 0.093 | 0.4831 | 0.4411 | 0.8185 |
| Model 2 | 0.7756 | 0.3077 | 0.1739 | 0.9088 | 0.1364 | 0.4938 | 0.4661 | 0.8256 |
| Model 3 | 0.8051 | 0.5217 | 0.2727 | 0.9152 | 0.3556 | 0.5377 | 0.5680 | 0.8457 |
| Model 4 | 0.8057 | 0.5116 | 0.1111 | 0.9277 | 0.3750 | 0.5231 | 0.5423 | 0.8511 |
| Model 5 | 0.7329 | 0.2105 | 0.1395 | 0.9152 | 0.1429 | 0.5315 | 0.4452 | 0.8219 |
| Model 6 | 0.7367 | 0.4000 | 0.1778 | 0.8918 | 0.2703 | 0.5231 | 0.4999 | 0.8104 |
| Model 7 | 0.7216 | 0.3333 | 0.2034 | 0.9124 | 0.1333 | 0.4981 | 0.4670 | 0.8171 |
| Model 8 | 0.7700 | 0.4583 | 0.1951 | 0.9165 | 0.3043 | 0.5214 | 0.5277 | 0.8355 |

*Precision-Recall Curve*

To further evaluate the trade-off between precision and recall, Precision-Recall (PR) curves were generated. Across all models, PR curves for the Male Adult class displayed high stability, maintaining both high precision and recall. Conversely, PR curves for Female Teenager and Male Child classes dropped sharply, indicating rapid degradation in precision with increasing recall. This behavior points to classification boundaries that are not well-separated for these classes.

Among the ResNet50 based models, Model 3 (ResNet50 + GAP + SGD) achieved the highest macro-average PR-AUC score of 0.5774 (**Table 6**), followed closely by Model 4 (ResNet50 + Max Pooling + SGD) with a score of 0.5497. These results demonstrate the advantage of deep pre-trained architectures in extracting generalizable features, even under constrained visual conditions. Notably, Model 4 also delivered the highest test accuracy among all models (86.19%), indicating strong alignment between its PR performance and overall classification effectiveness.

However, Model 8 (Custom CNN + Max Pooling + SGD) offered a compelling trade-off, achieving a macro-average PR-AUC of 0.5283 while using less than half the trainable parameters of the ResNet50 models and converging in just 18 epochs. Its PR curves (**Figure 6c**), while slightly less smooth than those of Model 3 or 4 (**Figure 6a, Figure 6b**), showed more balanced behavior across several underrepresented classes, including Female Adult and Male Teenager. Although challenges remained for Female Teenager, Model 8's gradual PR curve decay suggested better generalization than other custom models.

While ResNet50 variants demonstrated superior absolute PR-AUC values, the custom architecture in Model 8 proved that lightweight networks can achieve competitive precision-recall performance when supported by appropriate pooling strategies and optimization techniques.

**TABLE 6 Macro Average PR-AUC Scores for Different Models**

| Model | Macro Average PR-AUC Score |
|---|---|
| Model 1 | 0.4672 |
| Model 2 | 0.5158 |
| Model 3 | 0.5774 |
| Model 4 | 0.5497 |
| Model 5 | 0.4549 |
| Model 6 | 0.5075 |
| Model 7 | 0.4738 |
| Model 8 | 0.5283 |

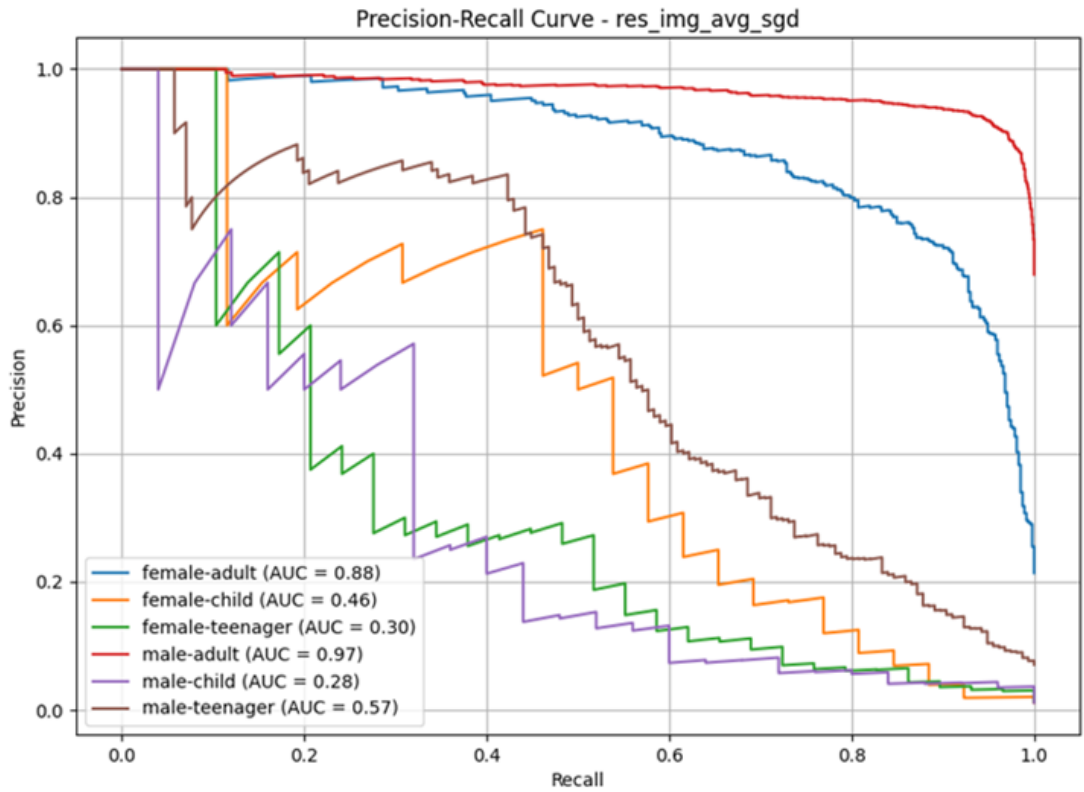

(a)

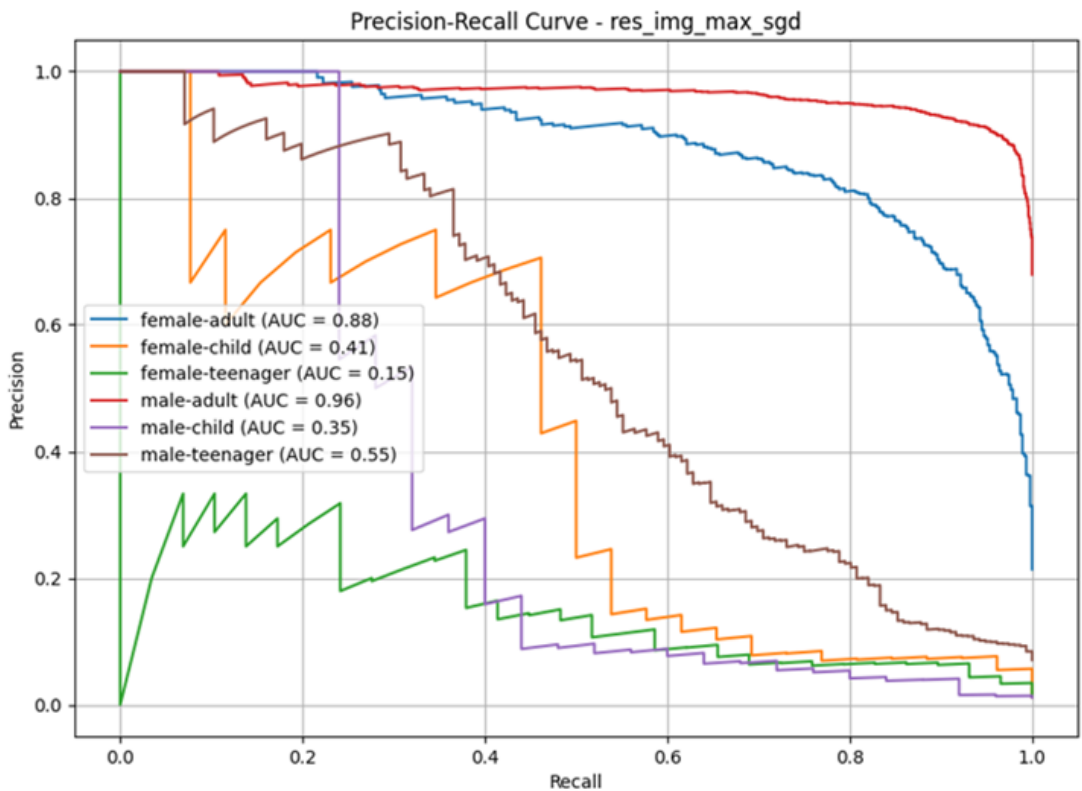

(b)

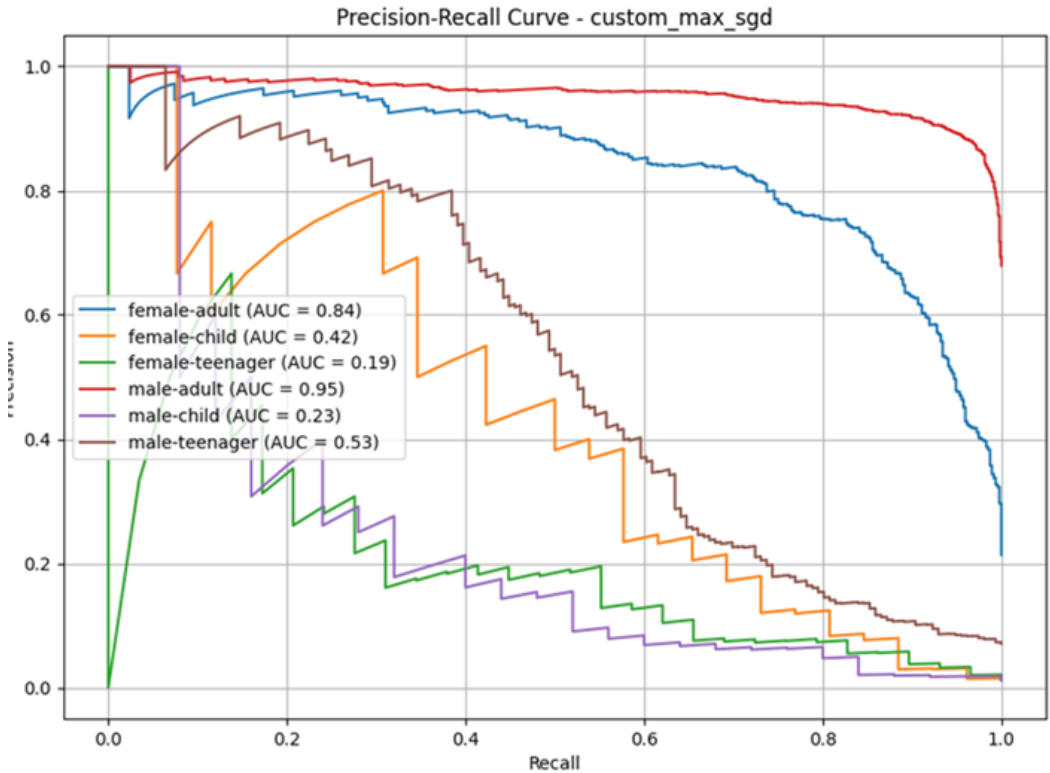

(c)
**Figure 6 Precision-Recall Curve (PR Curve) of (a) Model 3, (b) Model 4, (c) Model 8**

*Confusion Matrix*

To further examine misclassification patterns, confusion matrices were analyzed for all eight models. Consistently across models, the Male Adult class achieved the highest number of correct classifications, reflecting its dominance in the dataset and the distinctiveness of its visual features. In contrast, minority classes such as Female Teenager, Female Child, and Male Child experienced significantly higher misclassification rates. The most common errors included Female Teenager being misclassified as Female Adult and Male Child as Male Teenager, likely due to overlapping physical attributes such as height, posture, and clothing style, especially in the absence of facial cues from far-view images.

Despite applying data augmentation to balance the number of training samples per class, these confusion patterns persisted. This indicates that class imbalance alone was not the sole cause of poor performance. Rather it was compounded by visual similarity across adjacent demographic categories and limited discriminative features in low-resolution footage. Notably, Model 8 (Custom CNN) showed slight improvements in correctly identifying some minority classes, but overall, all models remained biased toward the better-represented categories. These observations highlight the need for additional strategies, such as incorporating pose or temporal features, to improve differentiation between visually similar pedestrian groups.

While data augmentation helped mitigate class imbalance to some extent, it could not fully offset the lack of inherent visual distinction in underrepresented classes. This points to the need for more sophisticated augmentation techniques or additional training data that introduces discriminative variation such as higher-resolution inputs, pose-based features, or multimodal information such as thermal imaging.

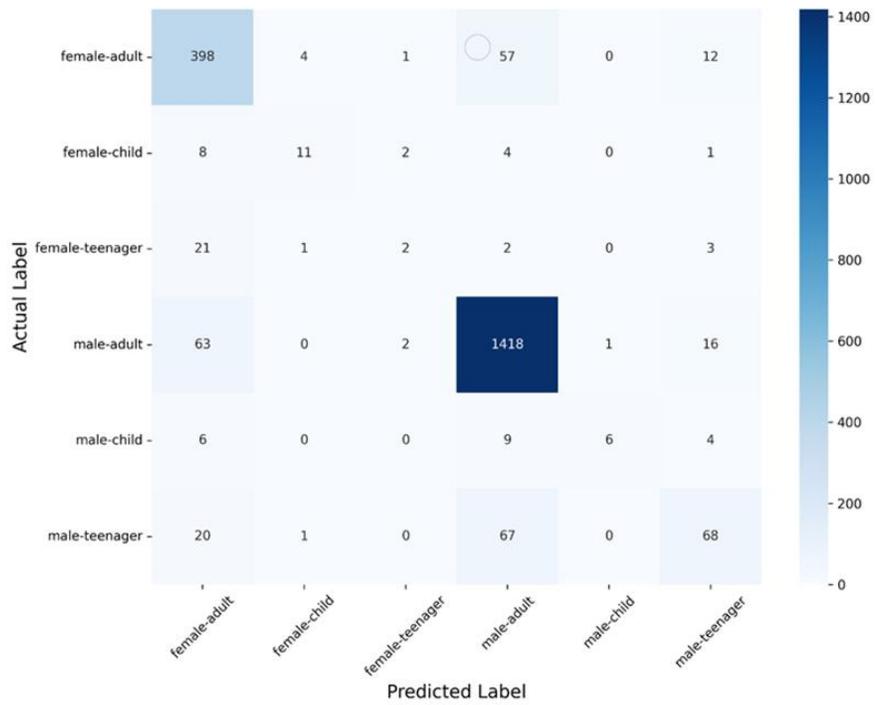

**Figure 7 Confusion Matrix of Model 4**

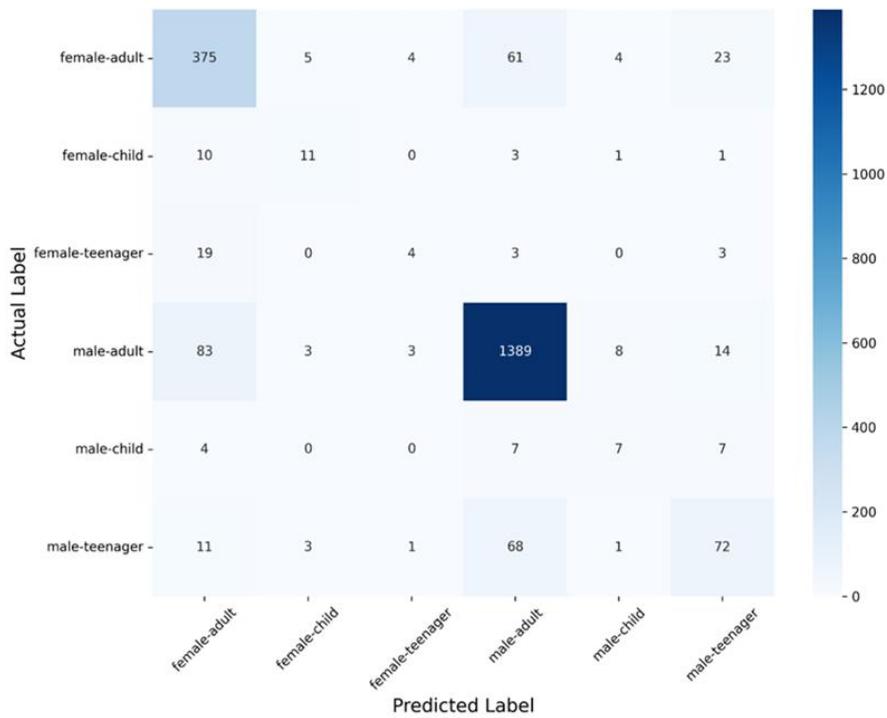

**Figure 8 Confusion Matrix of Model 8**

*Model Selection*

Given the small difference in performance between Model 4 (ResNet50 + MP + SGD) and Model 8 (Custom CNN + MP + SGD), but a significant reduction in parameters and training time for the latter, Model 8 is selected as the most suitable model for practical deployment. It balances strong accuracy (84.15%) with lower hardware demands and maintains relatively good recall and F1-scores for most classes, especially when computational constraints are a concern.

## CONCLUSIONS

This study presents a convolutional neural network–based framework for classifying pedestrian age group and gender using far-view traffic surveillance footage in complex, mixed-traffic environments. The classification task is structured as a single six-class problem, allowing the model to jointly infer age and gender by focusing on body-based visual cues such as posture and clothing style, rather than relying on facial features. This design choice enables the model to perform well even when image resolution is low, or faces are occluded. Among the eight trained models, the custom CNN with Max Pooling and SGD optimizer achieved strong performance (84.15% test accuracy) while maintaining high computational efficiency, making it a practical solution for resource-constrained deployments. In contrast, while ResNet50 variants slightly outperformed in accuracy, they required significantly more parameters and training time.

The system offers a scalable and cost-effective alternative to expensive sensors or facial recognition tools, making it particularly useful for cities with limited budgets and high pedestrian volumes. It can be implemented using existing traffic cameras without requiring high-resolution imaging, enabling deployment in resource-constrained environments. Transportation agencies can apply these insights to redesign intersections, introducing extended signal phases, raised medians, or refuge islands in areas with frequent child or elderly crossings. Gender related data can guide safety interventions such as improved lighting, wider sidewalks, or targeted awareness campaigns in locations with high female pedestrian presence. Unlike manual surveys, the model offers continuous, automated monitoring with minimal operational overhead. It can also enhance microsimulation and agent-based models by integrating realistic demographic variability in pedestrian speed or behavior. Additionally, the demographic outputs may be integrated with trajectory data to analyze movement stability and behavioral irregularities, enabling deeper insights into pedestrian dynamics. In dynamic policy contexts, such as near schools or hospitals, real-time outputs can trigger adaptive signal control, increasing safety during peak movement hours for vulnerable groups.

## ACKNOWLEDGEMENTS
This research did not receive any external funding. Grammarly was used for grammar and writing standards.

## AUTHOR CONTRIBUTIONS
The authors confirm contribution to the paper as follows: study conception and design: S. S. Arif, N. Haque, M. A. Raihan, M. Hadiuzzaman; data collection: S. S. Arif; analysis and interpretation of results: S. S. Arif, N. Haque; draft manuscript preparation: M. M. Shahrier, M. A. Raihan. All authors reviewed the results and approved the final version of the manuscript.